\documentclass[11pt]{article}
\usepackage[numbers]{natbib}
\usepackage{macros/packages}
\usepackage{macros/editing-macros}
\usepackage[T1]{fontenc}    %
\usepackage{hyperref}       %
\usepackage{url}            %
\usepackage{booktabs}       %
\usepackage{amsfonts}       %
\usepackage{nicefrac}       %
\usepackage{microtype}      %
\usepackage{xcolor}         %
\usepackage{wrapfig}
\usepackage{subcaption}
\usepackage{tikz}
\usepackage{algorithm}
\usepackage{algorithmic}
\usepackage{adjustbox}
\usepackage{wrapfig}
\usepackage{subcaption}  %

\usetikzlibrary{%
  calc,
  fit,
  arrows,
  arrows.meta,
  positioning,
  decorations.pathreplacing,
  decorations.shapes,
}
\theoremstyle{definition}

\usepackage{ulem}

\renewcommand{\emph}[1]{\textit{#1}}

\begin{document}

\abovedisplayskip=8pt plus0pt minus3pt
\belowdisplayskip=8pt plus0pt minus3pt

\begin{center}
    {\huge 
        Data Mixture Optimization: 
        \vspace{0.3cm} \\
        A Multi-fidelity Multi-scale Bayesian Framework 
    } \\
  \vspace{.5cm} {\large Thomson Yen$^{*}$ ~~~~ Andrew Wei Tung Siah$^{*}$ ~~~~ Haozhe Chen   \\ Tianyi Peng  ~~~~ Daniel Guetta ~~~~  Hongseok Namkoong   } \\
  \vspace{.2cm}
  {\large Decision, Risk, and Operations Division, Columbia Business School} \\
     \vspace{.2cm}
   \texttt{\{ty2531, andrew.siah, haozhe.chen, tp2845,  crg2133, hongseok.namkoong\}@columbia.edu}
  \vspace{.2cm}
  \texttt{}
\end{center}

\begin{abstract}%
    Careful curation of data sources can significantly improve the performance of LLM pre-training, but
predominant approaches rely heavily on intuition or costly trial-and-error, making them difficult to generalize across different data domains and downstream tasks. 
Although scaling laws can provide a principled and general approach for data curation, standard deterministic extrapolation from small-scale experiments to larger scales requires strong assumptions on the reliability of such extrapolation, whose brittleness has been highlighted in prior works.
In this paper, we introduce a \emph{probabilistic extrapolation framework} for data mixture optimization that avoids rigid assumptions and explicitly models the uncertainty in performance across decision variables. 
We formulate data curation as a sequential decision-making problem---multi-fidelity, multi-scale Bayesian optimization---where $\{$data mixtures, model scale, training steps$\}$ are adaptively selected to balance training cost and potential information gain. 
Our framework naturally gives rise to algorithm prototypes that leverage noisy information from inexpensive experiments to systematically inform costly training decisions.
To accelerate methodological progress, we build a simulator based on 472 language model pre-training runs with varying data compositions from the SlimPajama dataset.  
We observe that even simple kernels and acquisition functions can enable principled decisions across training models from 20M to 1B parameters and achieve \textbf{2.6x} and \textbf{3.3x} speedups compared to multi-fidelity BO and random search baselines. Taken together, our framework underscores potential efficiency gains achievable by developing principled and transferable data mixture optimization methods. 
Our code is \href{https://github.com/namkoong-lab/data-recipes}{publicly available}.
\end{abstract}
\def\thefootnote{*}\footnotetext{Equal contribution}
\section{Introduction}

\begin{figure*}
    \centering

    \includegraphics[width=0.32\linewidth]{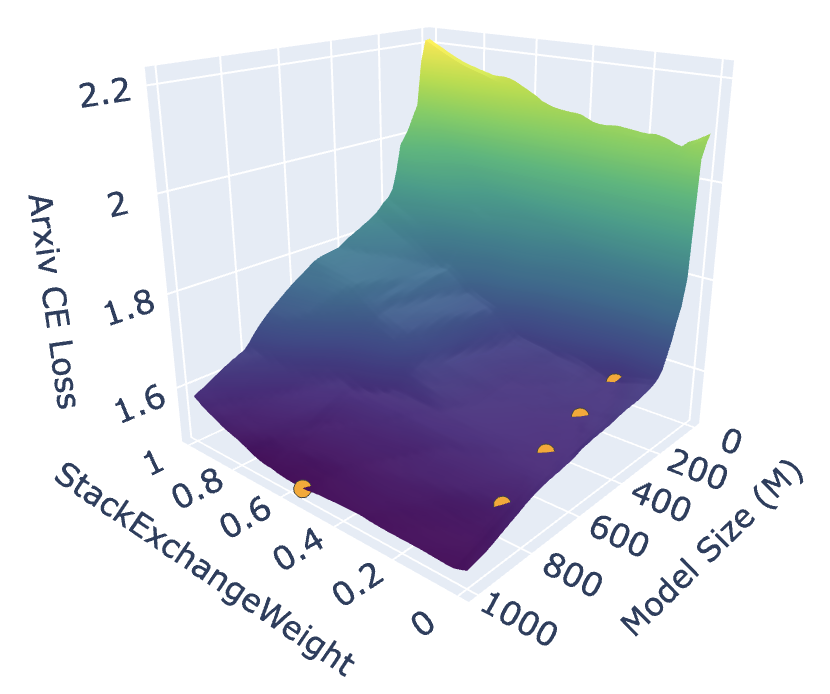}
    \includegraphics[width=0.66\linewidth]{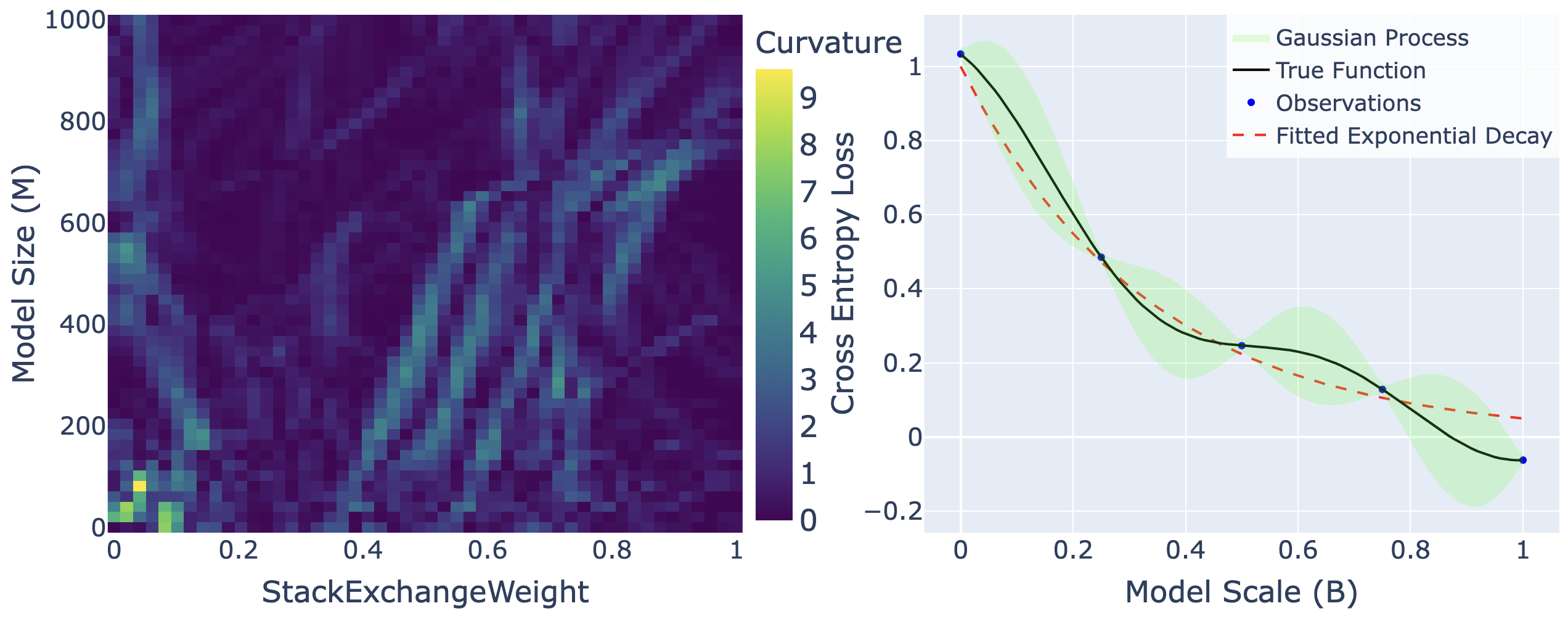}
    \caption{
        \textbf{Left:}
        The predicted validation cross-entropy loss on ArXiv data \citep{Shen_Slimpajama} as a function of data mixing coefficient and model sizes from a data-driven predictor on 472 runs (see details in Sec. \ref{sec:motivation}). 
        Notice the highly non-smooth geometry.
        Orange dots highlight the optimal data mixture proportion for each model scale.
        Note that they are not consistent across scales.
        \textbf{Middle: }
        The curvature at these points shows there are points of high irregularities, suggesting that the relationship between data mixture and model performance is unlikely to take a simple functional form. 
        \textbf{Right: }
        A demonstration showing how functional forms like exponential decay fitted on a small number of points would result in a high predictive error. 
        In contrast, a probabilistic model such as a Gaussian Process can capture uncertainty over the points. 
    }
    \label{fig:predictor}
\end{figure*}

\begin{figure*}[t]
    \centering
    \includegraphics[width=\textwidth]{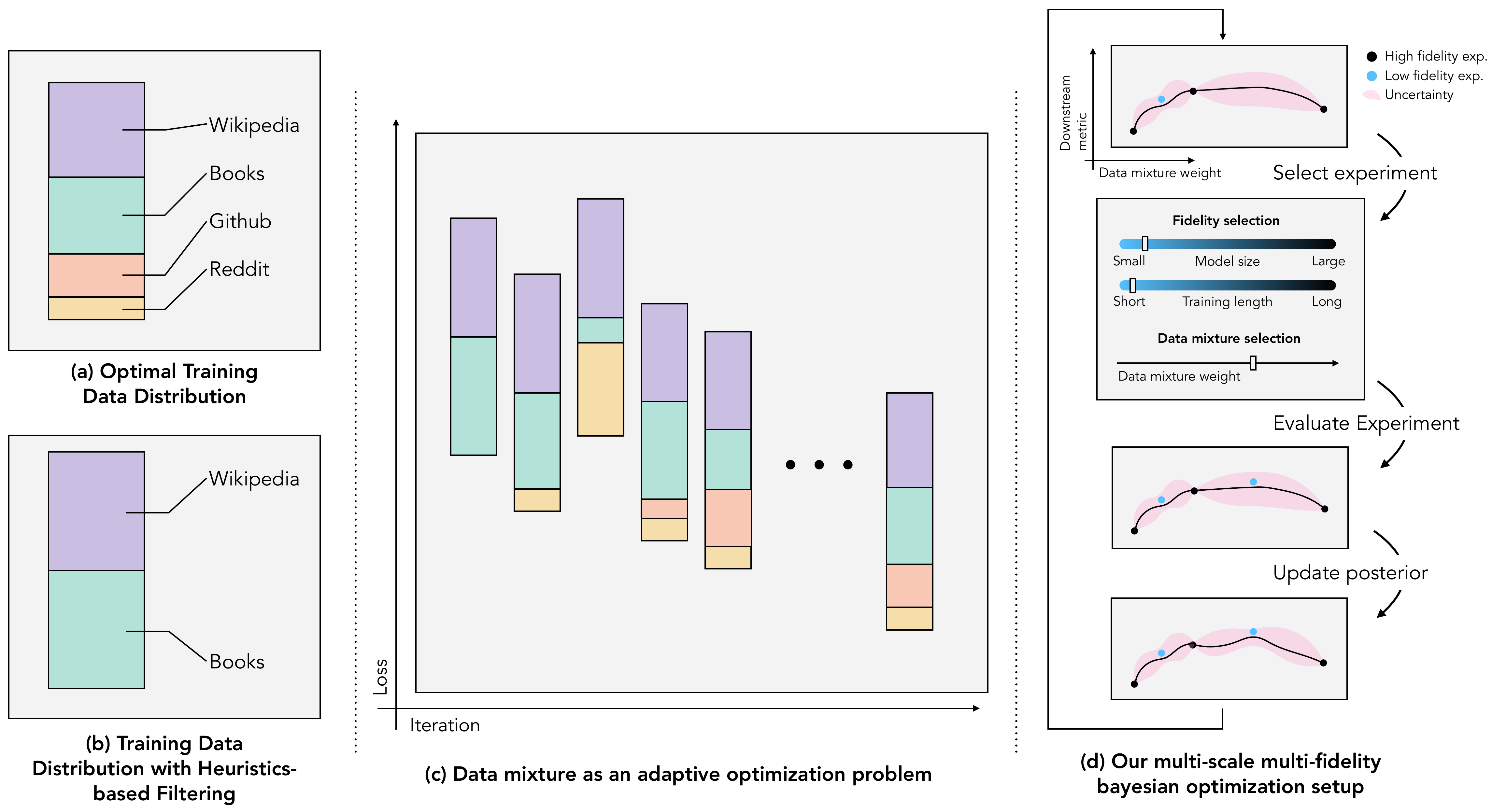}
    \caption{
        Our multi-fidelity multi-scale Bayesian optimization framework. (a) Given an unknown optimal training data distribution, (b) existing methods use heuristic-based filtering techniques to approximate the optimal distribution. (c) Our algorithm treats data mixture optimization as a Bayesian Optimization problem. (d) We explore data mixtures in a cost-aware fashion; when we test a new data mixture, we also choose the \emph{fidelity} of the observation we will observe. Larger models trained for more steps will result in \emph{high fidelity} observations, but be more expensive. Every point we observe updates our probabilistic belief of model performance over the data mixture, model size, and training steps space, which guides subsequent parameters. 
    }
    \label{fig:uq}
\end{figure*}
Data is the foundational infrastructure upon which all AI systems are built.
Scaling data has been a key driver of progress in machine learning, particularly in language model training \citep{Deng_Imagenet, Hoffmann_Chinchilla, Gadre_ScalingOverTrain}.
While this data-centric approach has led to impressive performance gains, it also incurs substantial computational and financial costs in training state-of-the-art models \citep{Hoffmann_Chinchilla, Luccioni_LLMCarbon}.
Beyond raw scale, the composition of training data has emerged as a critical factor in model performance \citep{Raffel_OnlineDataMixing, Goyal_DataFiltering}. 
For instance, TinyStories \citep{eldan2023tinystories} demonstrated that models with only 10 million parameters, when trained on a synthetic dataset, can generate coherent and consistent English text --- surpassing the capabilities of significantly larger models like GPT-2 Small (125 million parameters) \cite{radford2019_GPT2}.
Similarly, \citet{Li2024DCLM} showed that a carefully curated dataset enables training a 7-billion-parameter model comparable in performance to that of Mistral-7Bv0.3 and Llama 3 8B while using six times less compute. 
In practical scenarios where heterogeneous data sources are available, the choice of training mixture has been shown to significantly impact model performance.

This growing recognition of data composition's importance has led institutions to develop proprietary data mixtures based on domain expertise and empirical observations \citep{radford_CLIP, jiang2023mistral7b, openai_gpt4}.
Others have introduced heuristics, such as Wikipedia upsampling and perplexity-guided data selection, to refine training mixtures \citep{Tastu_PerplexityDataMixing, Blakeney_Upsampling}.
However, these ad hoc approaches are often tailored to specific training datasets and downstream tasks, and may not to transfer across domains and data types. 
For instance, when organizations in specialized sectors such as healthcare or finance seek to train custom language models on proprietary datasets, it remains unclear whether heuristics developed for public datasets are still effective. 
Given the substantial resources required for training high-performance language models, there is a pressing need for a principled framework to address data mixture optimization. 

Another line of research attempts to deterministically identify the functional relationship between data composition and model performance \citep{Ye_DataMixingLaw, Ge_BiMix, liu2025regmix}. 
Data Mixing Law \citep{Ye_DataMixingLaw}, for example, proposes fitting validation losses as an exponential function of linear combinations of data proportions. 
However, collecting sufficient data to fit parameters of such functions at the desired scale and training duration is often computationally prohibitive.
As a result, these methods inevitably rely on extrapolating the parameters learned from smaller models trained for fewer steps. 
Numerous studies have highlighted the brittleness of using such naive extrapolation to guide hyperparameter decisions \citep{Levine_DepthEfficiency, Ge_MuTransfer, jiang2025adaptive}. 
Notably, \citet{jiang2025adaptive} demonstrated that extrapolating validation losses based on small models often leads to inaccurate results. 
We reinforce this finding in our empirical study, observing that the optimal data mixing proportion does not remain constant across model scales (see Figure.~\ref{fig:predictor})

Existing approaches, whether based on heuristic trial-and-error or deterministic but unverifiable extrapolation, risk failing to identify the optimal data mixture as experiment scale or data domains change.
Moreover, they share another fundamental limitation: when additional computational budget becomes available for training new models, these approaches offer no clear guidance on how to select the next promising data mixture to experiment with. 
To address these challenges, we propose viewing the problem of curating the optimal data mixture as an adaptive optimization problem, where practitioners iteratively refine their mixing decisions based on empirical observations from prior experiments. 
This framework leverages the intuition that model performance exhibits local consistency across similar mixtures, training steps, and model scales, while avoiding rigid assumptions about the global structure of the performance landscape. 

Sequential optimization of data mixtures necessitates comprehending which data compositions suffer the highest uncertainty and sharpening beliefs on performance as more observations are gathered. In particular, good adaptive policies must distinguish between aleatoric and epistemic uncertainty: epistemic uncertainty can be reduced with more data, while aleatoric uncertainty is irreducible. Measurements must be planned to maximally reduce epistemic uncertainty on future runs by balancing exploration and exploitation.

We formulate this sequential optimization framework as a Bayesian optimization (BO) problem: we maintain probabilistic beliefs on the performance of various data mixtures and model scales, and use these beliefs to choose the next model scale to train, on what data mixture, and for how long. Once we fit and evaluate this new model, we use its performance to update our beliefs \citep{hutter2011, hutter2018, Frazier_BayesOptTutorial}. 

In traditional BO, the cost of each new observation is the same, and we aim to optimize an objective while observing the smallest number of points possible. Our setting is more complicated -- the cost of training a new model and observing its performance is affected by (1) the number of steps for which the model is trained and (2) the scale of the model (the number of parameters therein).

The number of steps for which a model is trained affects the \emph{quality} of the observation -- the more steps we use to train the model, the more accurately the results will reflect the utility of training on the data mixture in question. Previous work has handled this conundrum using so-called multi-fidelity Bayesian optimization, in which evaluations are `stopped early' during the training process if it becomes clear the information revealed during additional training steps will not be worth the expense
\citep{Swersky_FreezeThaw, Tobias_MFBO, Kandasamy_BOCA, Li_Hyperband}.

Our setting is distinguished by the second factor above -- we also want to use data gathered on smaller model scales to guide our search over parameters for larger models. 
We refer to this new setting as \textit{Multi-Fidelity Multi-Scale} (MFMS) Bayesian Optimization.
Importantly, varying model scale differs fundamentally from traditional fidelity dimensions like training steps. When training for $z$ steps, we naturally obtain observations for all intermediate steps up to $z$. In contrast, evaluating a model of size $m$ provides no inherent information about the performance of smaller or larger architectures. 
This raises interesting questions about how to appropriately treat and exploit this structure, opening new methodological directions for investigation.

Fortunately, unlike conventional hyperparameters such as learning rate or momentum, where optimal configurations exhibit complex scaling behavior across model sizes \citep{Ge_MuTransfer}, recent empirical evidence suggests that optimal data mixture compositions enjoy greater transferability from smaller to larger model architectures \citep{Ye_DataMixingLaw, Ge_BiMix}.
This transferability property enables the strategic use of smaller-scale evaluations to identify optimal data mixture configurations that remain effective at target model scales, substantially reducing the computational cost of the optimization process. 

The main contributions of the paper are as follows:
\begin{itemize}
    \item We propose a probabilistic extrapolation framework to address the problem of optimizing data mixture for training LLMs. 
    The framework avoids rigid assumptions on the functional dependence between decision variables and model performance by explicitly modeling the performance uncertainty, thereby highlighting the need to strategically experiment with selected data mixtures to minimize the uncertainty in the optimal data mixture at the desired scale.  
    \item Our framework gives rise to a \textit{Multi-Fidelity Multi-Scale Bayesian Optimization} problem, which provides a principled foundation for developing and evaluating transferable methods for optimizing data mixture. This novel setting contains a rich structure, which opens up interesting avenues for future methodological advancements, including batching strategies, asynchronous optimization, custom kernels, and look-ahead methods. 
    \item To spur methodological progress, we build an empirical testbed based on a simulator trained on 472 language model pre-training runs with varying data compositions from the SlimPajama dataset. We demonstrate the promises of MFMS Bayesian Optimization setting by benchmarking a simple Gaussian-process-based method against baselines such as Hyperband \cite{Li_Hyperband} and Random Search, which do not consider model scale as a decision variable. 
    We find that even this simple approach can better explore different data mixtures and model scales, and deliver the best terminal model (as measured by downstream task performance) at least \textbf{2.6x} faster than baselines (Section~\ref{sec:bayesopt_sec}). 
\end{itemize}
\section{Motivation for Multi-fidelity Multi-scale Framework}
\label{sec:motivation}

Deterministically extrapolating optimal data mixtures from small-scale experiments to large-scale models can be unreliable, but results from smaller models and earlier training steps still provide valuable information that can guide large-scale training decisions.
In this section, we experimentally validate two key premises of Multi-Fidelity Multi-Scale Bayesian Optimization: (1) smaller models can help predict the performance of larger models under various data mixtures, and (2) undertrained models trained for fewer steps over different data mixtures can inform the performance of fully-trained models.

To demonstrate the predictive utility of smaller-scale experiments, we train predictors that take language model training parameters --- $\{$data mixtures, model scale, training steps$\}$ --- as input and predict either validation losses or downstream task accuracies of the resulting language model. 
Our experiments show that incorporating results from smaller-scale experiments improves the predictors' ability to estimate model performance as a function of data mixture at larger scales. 

\emph{It is important to note that these predictors are distinct from the language models themselves.
The predictors take training hyperparameters as input and forecast the resulting language model's performance. }

\subsection{Collecting Predictor's Data Through LLM Pre-training Runs}
\label{subsec:predictor_data}
To collect data to train the predictors, we pretrained 472 language models using the OLMo 2 package \citep{olmo20242olmo2furious} and data from SlimPajama \citep{Shen_Slimpajama}, a deduplicated version of RedPajama \citep{Weber_RedPajama}.
SlimPajama contains seven data categories -- \textit{Wikipedia}, \textit{StackExchange}, \textit{Github}, \textit{ArXiv}, \textit{Book}, \textit{CommonCrawl}, and \textit{C4}. 
We used only data from the first five categories to pretrain the language models while holding out the data from \textit{CommonCrawl} and \textit{C4} to simulate data mixture optimization in out-of-distribution settings. 
We measure the training losses and separate the validation losses of each of the seven categories. 
Furthermore, we evaluate the language models on three downstream tasks: \emph{HellaSwag}, \emph{PIQA}, and \emph{Arc Easy} \cite{Bisk2020PIQA, allenai:arc, zellers2019hellaswag}.

For each run, we randomly sample the data mixture proportions from a Dirichlet distribution to uniformly sample from the probability simplex and train the models for 196 training steps. 
Under this setup, we trained models ranging from 20M to 1B parameters. 
Additional details on the pretraining setup are provided in Appendix~\ref{appendix:pretraining}.

\subsection{Predictor Training}\label{subsec:predictor_training}

We train predictors using multilayer perceptrons (MLPs) consisting of three hidden layers with 64 hidden units each, ReLU activations, and dropout with a rate of 0.1, totaling approximately 5,000 parameters. The predictors accept three types of inputs: (1) the model size (number of parameters), (2) the number of training steps, and (3) the proportions of each of the five dataset categories used during pretraining, namely (\textit{Wikipedia}, \textit{StackExchange}, \textit{Github}, \textit{ArXiv}, and \textit{Book}).

For each pretrained language model, the predictor outputs predictions for multiple metrics: the training loss, validation losses across seven categories (\textit{Wikipedia}, \textit{StackExchange}, \textit{Github}, \textit{ArXiv}, \textit{Book}, and held-out datasets \textit{CommonCrawl} and \textit{C4}), as well as downstream task accuracies on three evaluation tasks: \emph{HellaSwag}, \emph{PIQA}, and \emph{Arc Easy} \cite{zellers2019hellaswag,Bisk2020PIQA,allenai:arc}. Thus, each language model corresponds to a single row in our predictor's dataset, comprising 9 inputs as described above and outputs spanning these 11 metrics.

We train the predictors to maximize the coefficient of determination ($R^2$), $R^2 = 1 - \frac{\sum_{i}(y_i - \hat{y}_i)^2}{\sum_{i}(y_i - \bar{y})^2}$, where $y_i$ is the true metric value for data point $i$, $\hat{y}_i$ is the predictor's estimate, and $\bar{y}$ is the mean of true values. Training of our predictor is conducted for 20 epochs using a batch size of 64, an Adam optimizer with a learning rate of 0.001 and weight decay of 0.01, and data normalization (standard scaling) applied to both inputs and outputs.

\subsection{Small Models Help Predict Larger Models Outcomes}\label{subsec:smallhelpslarge}
\setcounter{table}{2}
\begin{wrapfigure}{l}{0.53\textwidth}
\centering
\begin{tabular}{lccc|cc}
\toprule
\textbf{Dataset} & $E_1$ & $E_2$ & $E_3$ & $E_4$ & $E_5$ \\
\midrule
\text{Wikipedia}    & 0.75 & \textbf{0.96} & 0.94 & 0.73 & \textbf{0.88} \\
\text{ArXiv}        & 0.68 & \textbf{0.92} & 0.93 & 0.59 & \textbf{0.82} \\
\text{Github}       & 0.66 & 0.95 & \textbf{0.95} & 0.62 & \textbf{0.87} \\
\text{Book}         & 0.83 & \textbf{0.97} & \textbf{0.97} & 0.79 & \textbf{0.92} \\
\text{StackExchange} & 0.73 & \textbf{0.95} & \textbf{0.95} & 0.68 & \textbf{0.90} \\
\text{CommonCrawl}  & 0.84 & \textbf{0.98} & \textbf{0.98} & 0.81 & \textbf{0.94} \\
\text{C4}           & 0.86 & \textbf{0.99} & 0.98 & 0.82 & \textbf{0.95} \\
\text{ArcEasy}      & 0.92 & \textbf{0.94} & \textbf{0.94} & 0.88 & \textbf{0.90} \\
\text{HellaSwag}    & 0.97 & \textbf{0.98} & 0.97 & 0.94 & \textbf{0.96} \\
\text{PIQA}         & 0.94 & \textbf{0.96} & \textbf{0.96} & 0.90 & \textbf{0.93} \\
\bottomrule
\end{tabular}
\captionof{table}{
    \label{model_size_exp_result} $R^2$ values of the experiments listed in Table~\ref{model_size_exp_setup}, averaged over 3 random seeds. Notice that $E_2 > E_1$ and $E_5 > E_4$ -- our ability to predict the performance of larger models is considerably enhanced by insights from smaller models. Note also that $E_3 \approx E_2$; adding information about \emph{much} smaller models does not seem to help.
}
\vspace{-2em}
\end{wrapfigure}

We begin by investigating the extent to which smaller model runs can inform the dynamics of larger ones, as the literature on scaling laws  \cite{Gadre_ScalingOverTrain} would suggest. Table~\ref{model_size_exp_setup} details these experiments, and Table~\ref{model_size_exp_result} lists the results of the experiment.

\setcounter{table}{0}
\begin{table}[t]
    \centering
    \begin{minipage}{0.45\textwidth}
        \centering
    \begin{tabular}{lccc}
    \toprule
    \textbf{} & \textbf{Train} & \textbf{Test} \\
    \midrule
    $E_1$    & half of 1B runs & remaining 1B runs \\
    $E_2$        & half of 1B runs & remaining 1B runs \\
                & + 700M runs & \\
    $E_3$       & half of 1B runs & remaining 1B runs \\
                & + all smaller runs & \\
    $E_4$         & half of 700M runs& remaining 700M runs\\          
    $E_5$         & half of 700M runs& remaining 700M runs\\
                & + 500M runs & \\
    \bottomrule
    \end{tabular}
    \captionof{table}{\label{model_size_exp_setup} Model size experiments}
    \end{minipage}
    \hfill
    \begin{minipage}{0.45\textwidth}
        \centering
\begin{tabular}{lccc}
\toprule
\textbf{Dataset} & \textbf{$E_6$} & \textbf{$E_7$} & \textbf{$E_8$} \\
\midrule
$R^2$  & 0.69 & 0.77 & \textbf{0.82} \\
$R^2(\log)$  & 0.74 & 0.82 & \textbf{0.85} \\
\bottomrule
\end{tabular}
\captionof{table}
{\label{exp2_table_result}Notice the predictive power of our MLP is strongest when it is trained on many runs for fewer steps. Results averaged over 3 runs. This provides some support for our intuition that, given a fixed compute budget, it is better to have more runs with fewer training steps than fewer runs for a large number of training steps. }
    \end{minipage}
\end{table}

We note that, as expected, information garnered from training runs on \emph{smaller} models seems to considerably increase the accuracy of our predictions on \emph{larger} models, motivating our hope that a carefully crafted optimization algorithm can exploit the relationship.

Unsurprisingly, we note that $E_3 \approx E_2 > E_1$ and $E_5 > E_4$: the closer in scale the smaller models are to the larger model about which we wish to make a prediction, the more useful the information is. We, therefore, expect our MFMS BayesOpt algorithms to `step through' model scales, starting with small and cheap models to identify promising data mixtures, and then progressing to larger and larger models, all the while refining the data mixtures it considers optimal.

\subsection{Earlier Training Steps Help Predict Later Training Steps}\label{subsec:lowhelpshigh}

The second central premise of our approach is that, given a fixed compute budget, it can be better to attempt many runs for fewer training steps than fewer runs for a larger number of training steps.

To test this hypothesis, we carried out three additional experiments. In each of these experiments, we attempt to predict the final losses in 30\% of our model runs (evenly distributed across model sizes). The MLP for each of these experiments is trained on (1) a set of complete runs, one for each model size (2) a set of `truncated' runs, evenly distributed across model sizes. In $E_6$, we use 16 runs truncated at 196 training steps, in $E_7$, we use 22 runs truncated at 130 training steps, and in $E_8$, we use 32 runs truncated at 85 steps; thus, these experiments are trained on numbers generated with the \emph{same} FLOPS budget.

\section{Multi-Fidelity Multi-Scale Bayesian Optimization} 
Knowing that both smaller models and early stopping can provide valuable insights for optimizing data mixtures, practitioners face a critical dilemma:
Given a fixed computational budget, how should they allocate resources to train the best model?
Should they train many small models to explore different data mixtures, use early stopping at the target scale to abandon poor-performing data mixtures, or adopt a hybrid approach combining both strategies?
The \textit{Multi-Fidelity Multi-Scale} setting articulates this dilemma, and in this section, we provide a mathematical formulation of the problem.

We consider having access to a set of $n$ datasets $\mathcal{D} = \{D_1, D_2, \ldots, D_n\}$, and aim to train the best-performing language model, evaluated on a given metric, with $m^*$ parameters for $z^*$ training steps using $T$ datapoints. 
The key decision variables are the fractions of the data budget $T$ allocated to each dataset. 
Specifically, we sample $w_i T$ data points from $D_i$, where $\boldsymbol w = \{w_1, w_2, ..., w_n\} \in \Delta^n$ and $\Delta^n$ denotes $n$-dimensional probability simplex.
Let $\mu(\boldsymbol{w}, m, z)$ denote the performance of a model with $m$ parameters trained for $z$ training steps using dataset proportions $\boldsymbol{w}$. 

The model's performance is represented by an unknown function $\mu(\boldsymbol{w}, m^*, z^*)$, which maps a given data mixture $\boldsymbol{w}$, model size $m$, and training steps $z$ to an evaluation metric.
Our goal is to solve the optimization problem: 
\begin{eqnarray}
    \arg\max_{\boldsymbol{w}} \mu(\boldsymbol{w}, m^*, z^*)
\end{eqnarray}
where $m^*$ and $z^*$ represent the target model size and training steps.

We have a budget $B$ with which we can experiment using different values of $\boldsymbol{w}$, $m$, and $z$. Each evaluation of $\mu(\boldsymbol{w}, m, z)$ incurs a cost $c(m, z)$. $c(m, z)$ is typically an increasing function of $m$ and $z$, though our framework does not require this. 
Using $m^*$ parameters and $z^*$ steps every time we evaluate a new set of weights would quickly exhaust our budget. Instead, therefore, we might probe a particular set of weights on a smaller model with $m < m^*$ parameters, or with $z < z^*$ training steps --- while the resulting observation $\mu(\boldsymbol{w}, m, z)$ would be less informative than $\mu(\boldsymbol{w}, m^*, z^*)$, it would be considerably cheaper and still provide valuable information.

However, since $\mu(\cdot)$ is an unknown function, it is critical to address the uncertainty in its values.
Bayesian Optimization provides a natural framework that explicitly models uncertainty in $\mu(\cdot)$ and systematically refines estimates of $\mu(\cdot)$ through posterior updates, as one observes more evaluations at different input configurations.
This technique is called multi-fidelity Bayesian optimization.
See Figure~\ref{fig:uq} for a graphical representation of the setup. 

Traditional approaches to fidelity-aware Bayesian optimization primarily address scenarios in which the model architecture $m$ remains fixed and only the number of training steps $z$ is varied \citep{Swersky_FreezeThaw, Tobias_MFBO, Kandasamy_BOCA, Li_Hyperband}. We add a layer of complexity by also considering model scale. 
One might be tempted to regard model scale as merely an additional fidelity dimension, otherwise identical to training steps. 
However, we note that there is a fundamental distinction between the two dimensions:
In the course of evaluating a model trained for $z$ training steps, we must also evaluate that model for all steps  $z' < z$, whereas no such hierarchical relationship exists for evaluations across different model scales. 
This structural difference suggests promising avenues for novel methodological developments in multi-fidelity optimization theory, though such extensions lie beyond the scope of our present work.

\section{Experimental Setup} \label{sec:bayesopt_sec}

\subsection{Evaluation}

We propose \textit{Multi-Fidelity Multi-Scale Bayesian Optimization} as a natural framework to find optimal data mixture in practice. 
To demonstrate the potential of our framework, we use a simple Bayesian optimization algorithm based on Gaussian processes to optimize the data mixture.
However, training a new language model for each parameter that the algorithm proposes would be prohibitively expensive. 
Instead, therefore, we train a predictor as described in Section~\ref{subsec:predictor_training}, and use the predictor to benchmark our approach.
We train the predictor on 472 language model training runs described in Section~\ref{subsec:predictor_data}, where 422 of the runs are randomly selected as a training set and the remaining 50 runs as a validation set. 
The predictor achieves an $R^2 >= 0.95$ across all metrics.
\subsection{Multi-fidelity Multi-scale Gaussian Process (MFMS-GP)}
\label{subsec:mfms-gp}
\begin{algorithm}
\caption{Gaussian Process and EIpu}
\begin{algorithmic}[1]
\STATE \textbf{Input:} Probability space $\Delta^n$, model-scale space $\mathcal M$, training-step space $\mathcal Z$, and cost function $c(\cdot, \cdot)$
\STATE Initialize Gaussian Process (GP) surrogate model with three RBF kernels over $\Delta^n$, $\mathcal M$, and $\mathcal Z$ and a linear mean function
\STATE Randomly sample points from $\Delta^n$, $\mathcal M$, and $\mathcal Z$ to initialize hyperparameters of GP. 
\STATE Initialize history $\mathcal{H}$ with the randomly sampled points
\FOR{each optimization iteration}
    \FOR{each $(m, z) \in \mathcal M \times \mathcal Z$}
        \STATE Optimize the EI evaluated at these values of $m$ and $z$ with respect to $\boldsymbol{w}$ gradient descent
    \ENDFOR
    \STATE Select the next configuration $\lambda_{\text{next}} = (\boldsymbol{w}_{\text{next}}, m_{\text{next}}, z_{{\text{next}}})$ that maximizees the Expected Improvement per Unit (EIpu):
    \[
    \text{EIpu}(\lambda) = \frac{\text{EI}(\lambda)}{c(m, z)}
    \]
    \STATE Evaluate $\mu(\lambda_{\text{next}})$
    \STATE Store results in $\mathcal{H}$
    \STATE Update posterior of GP with $\mathcal{H}$
\ENDFOR
\STATE Return best $\boldsymbol{w}^* = \lambda^*[0]$ from configuration $\lambda^* = \arg\max_{\lambda \in \mathcal{H}} \mu(\lambda)$
\end{algorithmic}
\end{algorithm}

We implement a Gaussian Process (GP) surrogate model for our MFMS setting. 
The kernel of the GP is a product of three separate RBF kernels for the data proportion, the model scale, and the training step dimensions. 
To enable learning the positive correlation between model performance and both model scales and training steps, we use a linear mean function. 

For the acquisition function, we use Expected Improvement (EI). 
EI aims to quantify the expected gain over the current best-observed function value, $\text{EI}(\mathbf{x}) := \mathbb{E} \left[ \max(y^* - f(\mathbf{x}), 0) \right]$, where the expectation is taken over the posterior distribution predicted by the surrogate models, and \( y^* \) represents the current best-observed function value, given by \( y^* := f(\mathbf{x}_{\min}) \) \citep{Frazier_BayesOptTutorial}. 
The EI function quantifies the expected improvement in the objective value compared to the current best, thereby encouraging the selection of points that are likely to yield better performance. 

Equipped with EI, the usual Bayesian optimization approach proceeds by optimizing EI over the parameter space to find the most promising point to evaluate, using gradient-based methods such as L-BFGS-B \citep{L-BFGS-B}. 
However, motivated by the fact that the parameter space is discrete over parameter counts $(m)$ and training steps ($z$), we optimize EI over each unique tuple $(m, z)$.
Then, to account for the fact that evaluation for each tuple incurs varying costs $c(m, z)$, we chose to evaluate the point that has the greatest EI per unit cost (EIpu) \citep{DBLP:journals/corr/abs-2003-10870}.

To initiate the hyperparameters of MFMS-GP, we randomly select $20$ configurations up to training step $z=9$ to fit the kernel and mean functions' parameters using the Adam optimizer \citep{Kingma2014AdamAM}.  
We provide additional details of our implementation in Appendix~\ref{appendix:bopt}.

\subsection{Baselines: Multi-fidelity Bayesian Optimization}

\begin{algorithm}
\caption{Hyperband with Random Forest, EI}
\begin{algorithmic}[1]
\STATE \textbf{Input:} Probability space $\Delta^n$, training-step space $\mathcal Z$, target model scale $m^*$, and cost function $c(m^*, \cdot)$
\STATE Initialize random forest surrogate model $\text{RF}$
\STATE Set initial design with Random Sampling
\STATE Initialize history $\mathcal{H} = \emptyset$
\FOR{each Hyperband iteration}
    \STATE Split the total computation budget into $s$ brackets
    \FOR{each bracket $s_i$}
        \STATE Generate initial configurations $\boldsymbol{w}_1, \dots, \boldsymbol{w}_n$ at lowest fidelity $z=1$
        \FOR{each fidelity $z$ from $1$ to $z^*$}
            \STATE Evaluate configurations $\boldsymbol{w}_i$ at fidelity $z$
            \STATE Store results in $\mathcal{H}$
            \STATE Fit $\text{RF}$ on $\mathcal{H}$
            \STATE Select next $\lambda_{\text{next}}$ using Expected Improvement
            \STATE Update $\mathcal{H}$ with new evaluations
        \ENDFOR
    \ENDFOR
\ENDFOR
\STATE Return best $\boldsymbol{w}^* = \lambda^*[0]$ from configuration $\lambda^* = \arg\max_{\lambda \in \mathcal{H}} \mu(\lambda)$
\end{algorithmic}
\end{algorithm}

As a baseline for multi-fidelity Bayesian optimization, we use Hyperband implemented by SMAC: Sequential Model-Based Optimization for General Algorithm Configuration \citep{lindauer-jmlr22a}.
This multi-fidelity Bayesian optimization uses a random forest as a surrogate model, expected improvement as the acquisition function, and uses Hyperband \citep{li2018hyperbandnovelbanditbasedapproach}, which is an early stopping technique that focuses on efficiently evaluating multiple parameter configurations by progressively eliminating poorly performing candidates, and exploring many combinations with fewer resources. 
Since the multi-fidelity framework does not offer a straightforward way to incorporate the additional dimension of model scale, throughout the optimization, we fix the number of the model's parameters to the target model scale $m^*$.  

\subsection{Baselines: Random Search}

For a baseline that does not consider utilizing the fidelity dimension, we consider random search.
Random search selects data proportions that are uniformly drawn from our data proportion space. 
We then run it against the largest model size and training steps. 

\section{Results}

In this section, we demonstrate the effectiveness of our MFMS-BO framework by benchmarking the simple MFMS-GP method against baselines that do not leverage smaller model scales. 

For each of the algorithms evaluated, we run 5 experiments over different seeds, and show the one standard deviation bound with shaded regions. 
The number of evaluations (the x-axis) is in terms of training FLOPS needed to train one 1B model at 100 training steps. 
As an example, for Random Search, the 1200 evaluation budget would allow sampling 6 full runs, where each run involves 20,000 training steps. 
The MFMS-GP method requires tuning its hyperparameters. 
As described in Section~\ref{subsec:mfms-gp}, we randomly select a few configurations to tune these parameters, and the cost of evaluating these configurations is accounted for. 

Since MFMS-GP relies on GP posterior and potentially noisy EI optimizations to select model scales and training steps, it may take a while to sample points at the target scale and fidelity. 
Therefore, we add an additional plot, MFMS-GP full-scale, that shows the performance one would have gotten if one takes the best configuration MFMS-GP has observed, and simply sets the model scale and training steps to the target $m^*$ and $z^*$. 
The additional computational cost of training at full scale is taken into account in the figures. 

\begin{figure*}[th!]
    \centering
    \includegraphics[width=\textwidth]{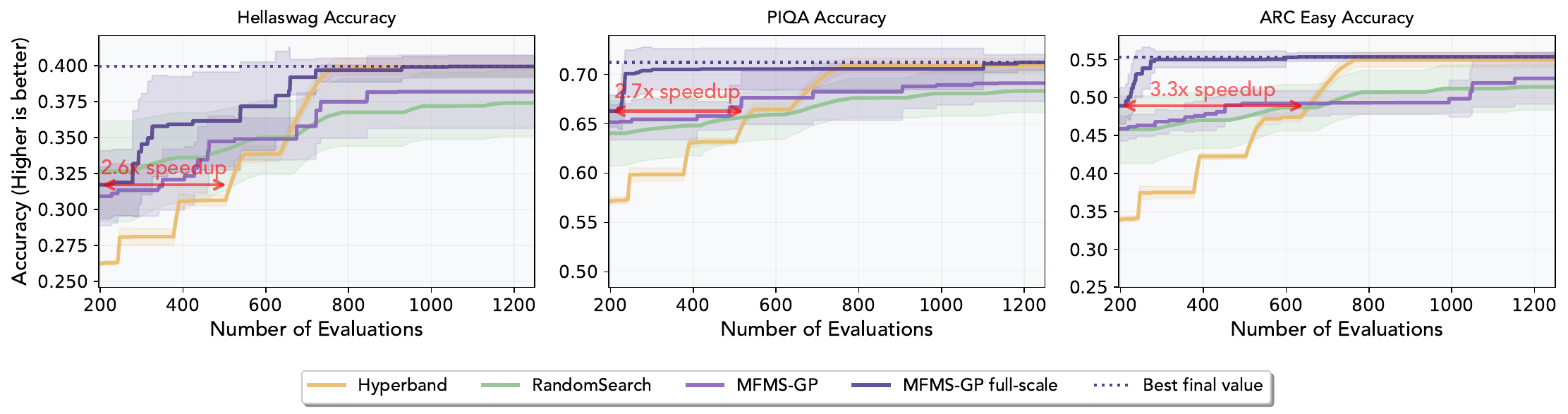}
    \caption{On maximizing accuracy in the downstream tasks, our multi-scale multi-fidelity approach achieves more than 2.6x speedup and finds the best configuration the fastest.}
    \label{fig:bayesopt1}
\end{figure*}

\begin{figure*}[th!]
    \centering
    \includegraphics[width=\textwidth]{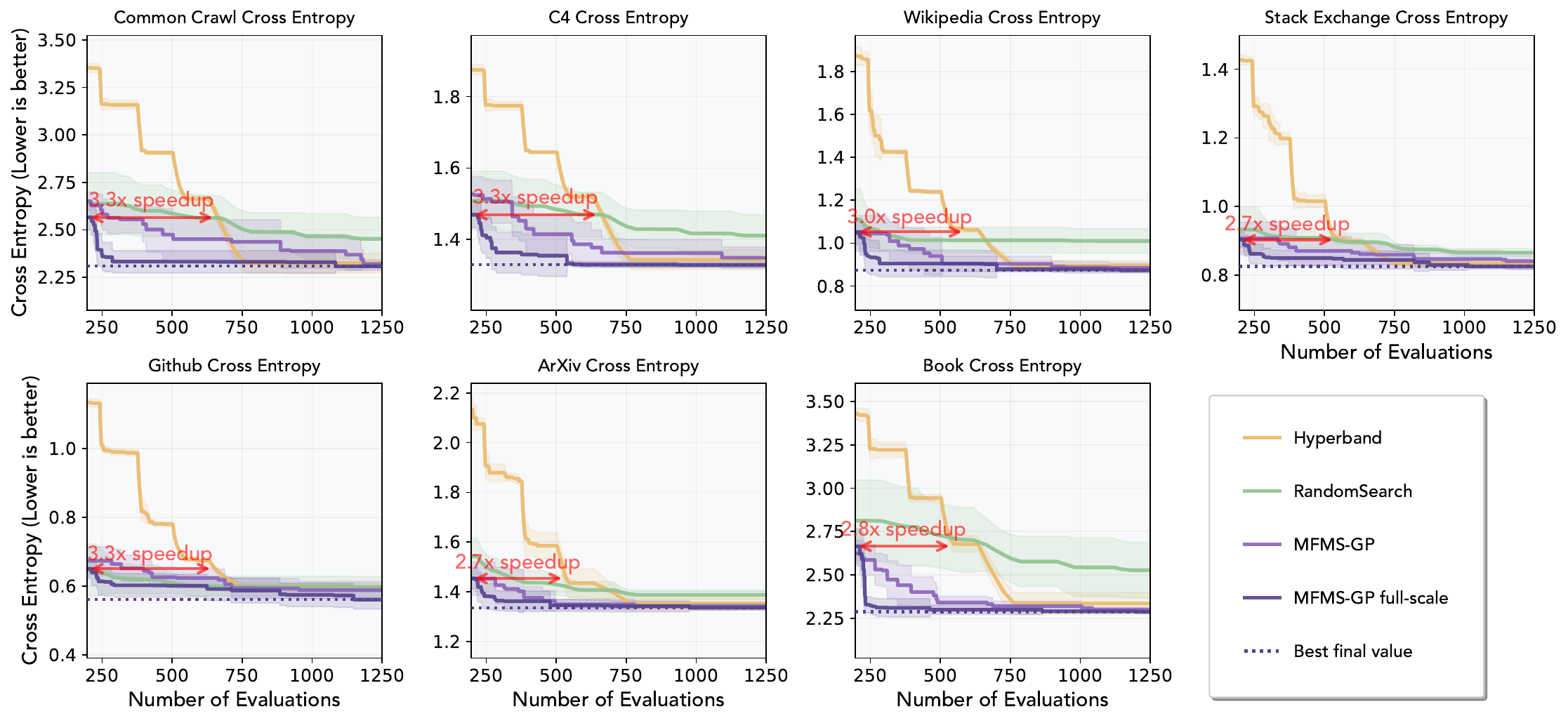}
    \caption{On minimizing the validation cross-entropy losses, our multi-scale multi-fidelity approach achieves more than 2.7x speedup and finds the best configuration the fastest.}
    \label{fig:bayesopt2}
\end{figure*}

In Figures~\ref{fig:bayesopt1} and \ref{fig:bayesopt2}, we see that both of the plots for our MFMS-GP algorithm have a \textbf{2.6x to 3.3x} speedup in finding the configuration that achieves the highest accuracy. 
Observe that MFMS-GP rapidly explores different data mixtures and identifies promising ones earlier than Hyperband or random search. 
This is expected, since MFMS-GP can utilize inexpensive evaluations by training on smaller models and for fewer steps. 
These strong results --- achieved with an extremely simple GP-based method --- underscore the potential of the MFMS framework in efficiently gathering low-cost yet valuable information to guide large-scale language model training. 
We expect that thoughtful algorithmic improvements, such as better kernel design or acquisition functions, will lead to even more powerful methods. 
We leave these explorations of the new setting for future works. 
\section{Related Work}

\textbf{Data Mixtures.} Several approaches aim to move beyond heuristic methods for data mixture by leveraging algorithmic techniques. \citet{albalak2023efficientonlinedatamixing} propose an online data mixing strategy using a non-stochastic bandit algorithm to dynamically adjust data proportions during training, maximizing perplexity. DoReMi \cite{xie2023doremioptimizingdatamixtures} focuses on identifying and emphasizing the ``hardest'' datasets for a base model through distributionally robust language modeling to improve training efficiency. \citet{ge2025bimixbivariatedatamixing} models a joint scaling behavior of domain proportions and training steps; we push this further through modeling the model scale. \citet{goyal2024scalinglawsdatafiltering} delve into the quality-quantity tradeoff in data, exploring how data filtering and repetition affect model performance and introducing scaling laws that account for data utility decay. These works highlight the increasing interest in principled and adaptive methods for data mixture optimization, yet often focus on fixed model scales. 
In contrast, our multi-fidelity multi-scale approach considers the practical scenario where practitioners have a fixed budget to experiment with data mixture, and can exploit cheap information gathered from smaller models and early training steps.

\textbf{Scaling Laws.} Scaling laws provide crucial insights into the relationship between model size, training compute, and performance in large language models \citep{kaplan2020scalinglawsneurallanguage}. \citet{hoffmann2022trainingcomputeoptimallargelanguage} established foundational scaling laws demonstrating predictable performance improvements with increased compute, model parameters, and training data.  
\citet{muennighoff2023scalingdataconstrainedlanguagemodels} investigates the impact of data repetition in data-constrained scenarios, showing diminishing returns beyond a certain repetition threshold. 
\citet{ruan2024observationalscalinglawspredictability} propose observational scaling laws based on "principal capabilities" to explain and predict language model performance across diverse models and benchmarks. 
These scaling law studies motivate our framework by providing an empirical basis for extrapolating performance variations of different data mixtures across model scales. 

\textbf{Bayesian Optimization.} Data mixture optimization, like hyperparameter tuning, benefits from efficient search strategies. Approaches range from full configuration selection with methods like Bayesian Optimization (BO) to configuration evaluation, which employs early termination of unpromising runs. Early BO methods \citet{hutter2011} used Gaussian Processes (GPs) to model the relationship between hyperparameters and model performance. Subsequent works explored random forests \citep{lindauer-jmlr22a} and Parzen estimators \citep{bergstra2011} as surrogate models. 
Early stopping techniques like Hyperband \citep{li2018hyperbandnovelbanditbasedapproach} focus on efficiently evaluating multiple parameter configurations by progressively eliminating poorly performing candidates, and exploring many combinations with fewer resources. More recent methods like BOHB \citep{hutter2018} combine these ideas, leveraging the BO exploration of Parzen estimators with the multi-fidelity benefits of Hyperband. Our work, MFMS-BO, is the first to explore a multi-fidelity multi-scale approach for data mixture optimization.

\section{Conclusion and Future Work}

This work introduces a principled framework, multi-fidelity multi-scale Bayesian optimization, for optimizing data mixture compositions in large language model training, a critical challenge in modern AI system development. 
Our framework unifies recent advances in predicting optimal data mixtures across scales with classical multi-fidelity Bayesian optimization techniques. 
Based on this unified framework, we implemented the Gaussian process using the RBF kernels and expected-improvement-per-unit acquisition function to balance the information gain and the cost of exploring new points in the functional landscape. 
We find that the method achieves optimal downstream task performance 2.6 times faster than traditional multi-fidelity approaches by strategically exploring the joint space of data mixtures and model scales.

In addition, we empirically demonstrate two key insights that inform future efficient optimization of data mixtures. 
First, our analysis reveals that training runs on smaller models (below 500M parameters) provide valuable predictive signals for optimizing larger architectures (1B parameters). 
Second, we establish that partial training runs can effectively inform full-scale training decisions. Specifically, our results show that a combination of full and partial training runs (e.g., 5 complete and 10 half-length runs) yields better predictive utility than an equal-compute allocation of full training runs alone (e.g., 10 complete runs).

Several promising directions emerge for future research. 
First, extending our framework to more settings, such as language model fine-tuning, data filtering, and more diverse collections of datasets, would help validate the framework's generalizability across different data mixing scenarios. 
From a methodological perspective, incorporating domain knowledge about the positive correlation between model performance and both parameter count and training duration could enhance the Gaussian process kernel design. 
Exploring alternative acquisition functions, such as knowledge gradient \citep{poloczek2016multiinformationsourceoptimization, wu2019mfbo}, could further improve efficiency in navigating the optimization landscape. 
The fundamental differences between model scale and training steps as fidelity dimensions also require further investigation to refine their appropriate treatment within the framework. 
Other realistic considerations could also enhance the practical benefits of the framework. 
One can consider developing algorithms that support batch evaluations and asynchronous updates to allow for more efficient parallel exploration. 
Addressing these challenges will further strengthen the framework and its applicability in large-scale language model training.

\newpage

\bibliographystyle{unsrtnat}

\setlength{\bibsep}{.7em}

\bibliography{bib.bib}

\begin{thebibliography}{49}
\providecommand{\natexlab}[1]{#1}
\providecommand{\url}[1]{\texttt{#1}}
\expandafter\ifx\csname urlstyle\endcsname\relax
  \providecommand{\doi}[1]{doi: #1}\else
  \providecommand{\doi}{doi: \begingroup \urlstyle{rm}\Url}\fi

\bibitem[Shen et~al.(2024)Shen, Tao, Ma, Neiswanger, Liu, Wang, Tan, Hestness,
  Vassilieva, Soboleva, and Xing]{Shen_Slimpajama}
Zhiqiang Shen, Tianhua Tao, Liqun Ma, Willie Neiswanger, Zhengzhong Liu, Hongyi
  Wang, Bowen Tan, Joel Hestness, Natalia Vassilieva, Daria Soboleva, and Eric
  Xing.
\newblock Slimpajama-dc: Understanding data combinations for llm training,
  2024.
\newblock URL \url{https://arxiv.org/abs/2309.10818}.

\bibitem[Deng et~al.(2009)Deng, Dong, Socher, Li, Li, and
  Fei-Fei]{Deng_Imagenet}
Jia Deng, Wei Dong, Richard Socher, Li-Jia Li, Kai Li, and Li~Fei-Fei.
\newblock Imagenet: A large-scale hierarchical image database.
\newblock In \emph{2009 IEEE Conference on Computer Vision and Pattern
  Recognition}, pages 248--255, 2009.
\newblock \doi{10.1109/CVPR.2009.5206848}.

\bibitem[Hoffmann et~al.(2022{\natexlab{a}})Hoffmann, Borgeaud, Mensch,
  Buchatskaya, Cai, Rutherford, de~Las~Casas, Hendricks, Welbl, Clark,
  Hennigan, Noland, Millican, van~den Driessche, Damoc, Guy, Osindero,
  Simonyan, Elsen, Rae, Vinyals, and Sifre]{Hoffmann_Chinchilla}
Jordan Hoffmann, Sebastian Borgeaud, Arthur Mensch, Elena Buchatskaya, Trevor
  Cai, Eliza Rutherford, Diego de~Las~Casas, Lisa~Anne Hendricks, Johannes
  Welbl, Aidan Clark, Tom Hennigan, Eric Noland, Katie Millican, George van~den
  Driessche, Bogdan Damoc, Aurelia Guy, Simon Osindero, Karen Simonyan, Erich
  Elsen, Jack~W. Rae, Oriol Vinyals, and Laurent Sifre.
\newblock Training compute-optimal large language models, 2022{\natexlab{a}}.
\newblock URL \url{https://arxiv.org/abs/2203.15556}.

\bibitem[Gadre et~al.(2024)Gadre, Smyrnis, Shankar, Gururangan, Wortsman, Shao,
  Mercat, Fang, Li, Keh, Xin, Nezhurina, Vasiljevic, Jitsev, Soldaini, Dimakis,
  Ilharco, Koh, Song, Kollar, Carmon, Dave, Heckel, Muennighoff, and
  Schmidt]{Gadre_ScalingOverTrain}
Samir~Yitzhak Gadre, Georgios Smyrnis, Vaishaal Shankar, Suchin Gururangan,
  Mitchell Wortsman, Rulin Shao, Jean Mercat, Alex Fang, Jeffrey Li, Sedrick
  Keh, Rui Xin, Marianna Nezhurina, Igor Vasiljevic, Jenia Jitsev, Luca
  Soldaini, Alexandros~G. Dimakis, Gabriel Ilharco, Pang~Wei Koh, Shuran Song,
  Thomas Kollar, Yair Carmon, Achal Dave, Reinhard Heckel, Niklas Muennighoff,
  and Ludwig Schmidt.
\newblock Language models scale reliably with over-training and on downstream
  tasks, 2024.
\newblock URL \url{https://arxiv.org/abs/2403.08540}.

\bibitem[Luccioni et~al.(2023)Luccioni, Viguier, and
  Ligozat]{Luccioni_LLMCarbon}
Alexandra~Sasha Luccioni, Sylvain Viguier, and Anne-Laure Ligozat.
\newblock Estimating the carbon footprint of bloom, a 176b parameter language
  model.
\newblock \emph{Journal of Machine Learning Research}, 24\penalty0
  (253):\penalty0 1--15, 2023.
\newblock URL \url{http://jmlr.org/papers/v24/23-0069.html}.

\bibitem[Albalak et~al.(2023{\natexlab{a}})Albalak, Pan, Raffel, and
  Wang]{Raffel_OnlineDataMixing}
Alon Albalak, Liangming Pan, Colin Raffel, and William~Yang Wang.
\newblock Efficient online data mixing for language model pre-training,
  2023{\natexlab{a}}.
\newblock URL \url{https://arxiv.org/abs/2312.02406}.

\bibitem[Goyal et~al.(2024{\natexlab{a}})Goyal, Maini, Lipton, Raghunathan, and
  Kolter]{Goyal_DataFiltering}
Sachin Goyal, Pratyush Maini, Zachary~C. Lipton, Aditi Raghunathan, and J.~Zico
  Kolter.
\newblock Scaling laws for data filtering—data curation cannot be compute
  agnostic.
\newblock In \emph{2024 IEEE/CVF Conference on Computer Vision and Pattern
  Recognition (CVPR)}, pages 22702--22711, 2024{\natexlab{a}}.
\newblock \doi{10.1109/CVPR52733.2024.02142}.

\bibitem[Eldan and Li(2023)]{eldan2023tinystories}
Ronen Eldan and Yuanzhi Li.
\newblock Tinystories: How small can language models be and still speak
  coherent english?, 2023.
\newblock URL \url{https://arxiv.org/abs/2305.07759}.

\bibitem[Radford et~al.(2019)Radford, Wu, Child, Luan, Amodei, and
  Sutskever]{radford2019_GPT2}
Alec Radford, Jeff Wu, Rewon Child, David Luan, Dario Amodei, and Ilya
  Sutskever.
\newblock Language models are unsupervised multitask learners.
\newblock 2019.

\bibitem[Li et~al.(2024)Li, Fang, Smyrnis, Ivgi, Jordan, Gadre, Bansal, Guha,
  Keh, Arora, Garg, Xin, Muennighoff, Heckel, Mercat, Chen, Gururangan,
  Wortsman, Albalak, Bitton, Nezhurina, Abbas, Hsieh, Ghosh, Gardner, Kilian,
  Zhang, Shao, Pratt, Sanyal, Ilharco, Daras, Marathe, Gokaslan, Zhang, Chandu,
  Nguyen, Vasiljevic, Kakade, Song, Sanghavi, Faghri, Oh, Zettlemoyer, Lo,
  El-Nouby, Pouransari, Toshev, Wang, Groeneveld, Soldaini, Koh, Jitsev,
  Kollar, Dimakis, Carmon, Dave, Schmidt, and Shankar]{Li2024DCLM}
Jeffrey Li, Alex Fang, Georgios Smyrnis, Maor Ivgi, Matt Jordan, Samir~Yitzhak
  Gadre, Hritik Bansal, Etash~Kumar Guha, Sedrick Keh, Kushal Arora, Saurabh
  Garg, Rui Xin, Niklas Muennighoff, Reinhard Heckel, Jean Mercat, Mayee~F
  Chen, Suchin Gururangan, Mitchell Wortsman, Alon Albalak, Yonatan Bitton,
  Marianna Nezhurina, Amro Kamal~Mohamed Abbas, Cheng-Yu Hsieh, Dhruba Ghosh,
  Joshua~P Gardner, Maciej Kilian, Hanlin Zhang, Rulin Shao, Sarah~M Pratt,
  Sunny Sanyal, Gabriel Ilharco, Giannis Daras, Kalyani Marathe, Aaron
  Gokaslan, Jieyu Zhang, Khyathi Chandu, Thao Nguyen, Igor Vasiljevic, Sham~M.
  Kakade, Shuran Song, Sujay Sanghavi, Fartash Faghri, Sewoong Oh, Luke
  Zettlemoyer, Kyle Lo, Alaaeldin El-Nouby, Hadi Pouransari, Alexander~T
  Toshev, Stephanie Wang, Dirk Groeneveld, Luca Soldaini, Pang~Wei Koh, Jenia
  Jitsev, Thomas Kollar, Alex Dimakis, Yair Carmon, Achal Dave, Ludwig Schmidt,
  and Vaishaal Shankar.
\newblock Datacomp-{LM}: In search of the next generation of training sets for
  language models.
\newblock In \emph{The Thirty-eight Conference on Neural Information Processing
  Systems Datasets and Benchmarks Track}, 2024.
\newblock URL \url{https://openreview.net/forum?id=CNWdWn47IE}.

\bibitem[Radford et~al.(2021)Radford, Kim, Hallacy, Ramesh, Goh, Agarwal,
  Sastry, Askell, Mishkin, Clark, Krueger, and Sutskever]{radford_CLIP}
Alec Radford, Jong~Wook Kim, Chris Hallacy, Aditya Ramesh, Gabriel Goh,
  Sandhini Agarwal, Girish Sastry, Amanda Askell, Pamela Mishkin, Jack Clark,
  Gretchen Krueger, and Ilya Sutskever.
\newblock Learning transferable visual models from natural language
  supervision, 2021.
\newblock URL \url{https://arxiv.org/abs/2103.00020}.

\bibitem[Jiang et~al.(2023)Jiang, Sablayrolles, Mensch, Bamford, Chaplot,
  de~las Casas, Bressand, Lengyel, Lample, Saulnier, Lavaud, Lachaux, Stock,
  Scao, Lavril, Wang, Lacroix, and Sayed]{jiang2023mistral7b}
Albert~Q. Jiang, Alexandre Sablayrolles, Arthur Mensch, Chris Bamford,
  Devendra~Singh Chaplot, Diego de~las Casas, Florian Bressand, Gianna Lengyel,
  Guillaume Lample, Lucile Saulnier, Lélio~Renard Lavaud, Marie-Anne Lachaux,
  Pierre Stock, Teven~Le Scao, Thibaut Lavril, Thomas Wang, Timothée Lacroix,
  and William~El Sayed.
\newblock Mistral 7b, 2023.
\newblock URL \url{https://arxiv.org/abs/2310.06825}.

\bibitem[OpenAI(2024)]{openai_gpt4}
OpenAI.
\newblock Gpt-4 technical report, 2024.
\newblock URL \url{https://arxiv.org/abs/2303.08774}.

\bibitem[Thrush et~al.(2024)Thrush, Potts, and
  Hashimoto]{Tastu_PerplexityDataMixing}
Tristan Thrush, Christopher Potts, and Tatsunori Hashimoto.
\newblock Improving pretraining data using perplexity correlations, 2024.
\newblock URL \url{https://arxiv.org/abs/2409.05816}.

\bibitem[Blakeney et~al.(2024)Blakeney, Paul, Larsen, Owen, and
  Frankle]{Blakeney_Upsampling}
Cody Blakeney, Mansheej Paul, Brett~W. Larsen, Sean Owen, and Jonathan Frankle.
\newblock Does your data spark joy? performance gains from domain upsampling at
  the end of training, 2024.
\newblock URL \url{https://arxiv.org/abs/2406.03476}.

\bibitem[Ye et~al.(2024)Ye, Liu, Sun, Zhou, Zhan, and Qiu]{Ye_DataMixingLaw}
Jiasheng Ye, Peiju Liu, Tianxiang Sun, Yunhua Zhou, Jun Zhan, and Xipeng Qiu.
\newblock Data mixing laws: Optimizing data mixtures by predicting language
  modeling performance, 2024.
\newblock URL \url{https://arxiv.org/abs/2403.16952}.

\bibitem[Ge et~al.(2025{\natexlab{a}})Ge, Ma, Chen, Li, and Ding]{Ge_BiMix}
Ce~Ge, Zhijian Ma, Daoyuan Chen, Yaliang Li, and Bolin Ding.
\newblock Bimix: A bivariate data mixing law for language model pretraining,
  2025{\natexlab{a}}.
\newblock URL \url{https://arxiv.org/abs/2405.14908}.

\bibitem[Liu et~al.(2025)Liu, Zheng, Muennighoff, Zeng, Dou, Pang, Jiang, and
  Lin]{liu2025regmix}
Qian Liu, Xiaosen Zheng, Niklas Muennighoff, Guangtao Zeng, Longxu Dou, Tianyu
  Pang, Jing Jiang, and Min Lin.
\newblock Regmix: Data mixture as regression for language model pre-training.
\newblock In \emph{The Thirteenth International Conference on Learning
  Representations}, 2025.
\newblock URL \url{https://openreview.net/forum?id=5BjQOUXq7i}.

\bibitem[Levine et~al.(2020)Levine, Wies, Sharir, Bata, and
  Shashua]{Levine_DepthEfficiency}
Yoav Levine, Noam Wies, Or~Sharir, Hofit Bata, and Amnon Shashua.
\newblock Limits to depth efficiencies of self-attention.
\newblock In H.~Larochelle, M.~Ranzato, R.~Hadsell, M.F. Balcan, and H.~Lin,
  editors, \emph{Advances in Neural Information Processing Systems}, volume~33,
  pages 22640--22651. Curran Associates, Inc., 2020.
\newblock URL
  \url{https://proceedings.neurips.cc/paper_files/paper/2020/file/ff4dfdf5904e920ce52b48c1cef97829-Paper.pdf}.

\bibitem[Yang et~al.(2021)Yang, Hu, Babuschkin, Sidor, Liu, Farhi, Ryder,
  Pachocki, Chen, and Gao]{Ge_MuTransfer}
Ge~Yang, Edward Hu, Igor Babuschkin, Szymon Sidor, Xiaodong Liu, David Farhi,
  Nick Ryder, Jakub Pachocki, Weizhu Chen, and Jianfeng Gao.
\newblock Tuning large neural networks via zero-shot hyperparameter transfer.
\newblock In M.~Ranzato, A.~Beygelzimer, Y.~Dauphin, P.S. Liang, and J.~Wortman
  Vaughan, editors, \emph{Advances in Neural Information Processing Systems},
  volume~34, pages 17084--17097. Curran Associates, Inc., 2021.
\newblock URL
  \url{https://proceedings.neurips.cc/paper_files/paper/2021/file/8df7c2e3c3c3be098ef7b382bd2c37ba-Paper.pdf}.

\bibitem[Jiang et~al.(2025)Jiang, Zhou, Feng, Malladi, and
  Kolter]{jiang2025adaptive}
Yiding Jiang, Allan Zhou, Zhili Feng, Sadhika Malladi, and J~Zico Kolter.
\newblock Adaptive data optimization: Dynamic sample selection with scaling
  laws.
\newblock In \emph{The Thirteenth International Conference on Learning
  Representations}, 2025.
\newblock URL \url{https://openreview.net/forum?id=aqok1UX7Z1}.

\bibitem[Hutter et~al.(2011)Hutter, Hoos, and Leyton-Brown]{hutter2011}
Frank Hutter, Holger~H. Hoos, and Kevin Leyton-Brown.
\newblock Sequential model-based optimization for general algorithm
  configuration.
\newblock In \emph{Proceedings of the 5th International Conference on Learning
  and Intelligent Optimization}, LION'05, page 507–523, Berlin, Heidelberg,
  2011. Springer-Verlag.
\newblock ISBN 9783642255656.
\newblock \doi{10.1007/978-3-642-25566-3_40}.
\newblock URL \url{https://doi.org/10.1007/978-3-642-25566-3_40}.

\bibitem[Falkner et~al.(2018)Falkner, Klein, and Hutter]{hutter2018}
Stefan Falkner, Aaron Klein, and Frank Hutter.
\newblock {BOHB:} robust and efficient hyperparameter optimization at scale.
\newblock \emph{CoRR}, abs/1807.01774, 2018.
\newblock URL \url{http://arxiv.org/abs/1807.01774}.

\bibitem[Frazier(2018)]{Frazier_BayesOptTutorial}
Peter~I. Frazier.
\newblock A tutorial on bayesian optimization, 2018.
\newblock URL \url{https://arxiv.org/abs/1807.02811}.

\bibitem[Swersky et~al.(2014)Swersky, Snoek, and Adams]{Swersky_FreezeThaw}
Kevin Swersky, Jasper Snoek, and Ryan~Prescott Adams.
\newblock Freeze-thaw bayesian optimization, 2014.
\newblock URL \url{https://arxiv.org/abs/1406.3896}.

\bibitem[Domhan et~al.(2015)Domhan, Springenberg, and Hutter]{Tobias_MFBO}
Tobias Domhan, Jost~Tobias Springenberg, and Frank Hutter.
\newblock Speeding up automatic hyperparameter optimization of deep neural
  networks by extrapolation of learning curves.
\newblock In \emph{Proceedings of the 24th International Conference on
  Artificial Intelligence}, IJCAI'15, page 3460–3468. AAAI Press, 2015.
\newblock ISBN 9781577357384.

\bibitem[Kandasamy et~al.(2017)Kandasamy, Dasarathy, Schneider, and
  P{\'o}czos]{Kandasamy_BOCA}
Kirthevasan Kandasamy, Gautam Dasarathy, Jeff Schneider, and Barnab{\'a}s
  P{\'o}czos.
\newblock Multi-fidelity {B}ayesian optimisation with continuous
  approximations.
\newblock In Doina Precup and Yee~Whye Teh, editors, \emph{Proceedings of the
  34th International Conference on Machine Learning}, volume~70 of
  \emph{Proceedings of Machine Learning Research}, pages 1799--1808. PMLR,
  06--11 Aug 2017.
\newblock URL \url{https://proceedings.mlr.press/v70/kandasamy17a.html}.

\bibitem[Li et~al.(2018{\natexlab{a}})Li, Jamieson, DeSalvo, Rostamizadeh, and
  Talwalkar]{Li_Hyperband}
Lisha Li, Kevin Jamieson, Giulia DeSalvo, Afshin Rostamizadeh, and Ameet
  Talwalkar.
\newblock Hyperband: A novel bandit-based approach to hyperparameter
  optimization.
\newblock \emph{Journal of Machine Learning Research}, 18\penalty0
  (185):\penalty0 1--52, 2018{\natexlab{a}}.
\newblock URL \url{http://jmlr.org/papers/v18/16-558.html}.

\bibitem[OLMo et~al.(2024)OLMo, Walsh, Soldaini, Groeneveld, Lo, Arora, Bhagia,
  Gu, Huang, Jordan, Lambert, Schwenk, Tafjord, Anderson, Atkinson, Brahman,
  Clark, Dasigi, Dziri, Guerquin, Ivison, Koh, Liu, Malik, Merrill, Miranda,
  Morrison, Murray, Nam, Pyatkin, Rangapur, Schmitz, Skjonsberg, Wadden,
  Wilhelm, Wilson, Zettlemoyer, Farhadi, Smith, and
  Hajishirzi]{olmo20242olmo2furious}
Team OLMo, Pete Walsh, Luca Soldaini, Dirk Groeneveld, Kyle Lo, Shane Arora,
  Akshita Bhagia, Yuling Gu, Shengyi Huang, Matt Jordan, Nathan Lambert, Dustin
  Schwenk, Oyvind Tafjord, Taira Anderson, David Atkinson, Faeze Brahman,
  Christopher Clark, Pradeep Dasigi, Nouha Dziri, Michal Guerquin, Hamish
  Ivison, Pang~Wei Koh, Jiacheng Liu, Saumya Malik, William Merrill, Lester
  James~V. Miranda, Jacob Morrison, Tyler Murray, Crystal Nam, Valentina
  Pyatkin, Aman Rangapur, Michael Schmitz, Sam Skjonsberg, David Wadden,
  Christopher Wilhelm, Michael Wilson, Luke Zettlemoyer, Ali Farhadi, Noah~A.
  Smith, and Hannaneh Hajishirzi.
\newblock 2 olmo 2 furious, 2024.
\newblock URL \url{https://arxiv.org/abs/2501.00656}.

\bibitem[Weber et~al.(2024)Weber, Fu, Anthony, Oren, Adams, Alexandrov, Lyu,
  Nguyen, Yao, Adams, Athiwaratkun, Chalamala, Chen, Ryabinin, Dao, Liang, Ré,
  Rish, and Zhang]{Weber_RedPajama}
Maurice Weber, Daniel Fu, Quentin Anthony, Yonatan Oren, Shane Adams, Anton
  Alexandrov, Xiaozhong Lyu, Huu Nguyen, Xiaozhe Yao, Virginia Adams, Ben
  Athiwaratkun, Rahul Chalamala, Kezhen Chen, Max Ryabinin, Tri Dao, Percy
  Liang, Christopher Ré, Irina Rish, and Ce~Zhang.
\newblock Redpajama: an open dataset for training large language models, 2024.
\newblock URL \url{https://arxiv.org/abs/2411.12372}.

\bibitem[Bisk et~al.(2020)Bisk, Zellers, Bras, Gao, and Choi]{Bisk2020PIQA}
Yonatan Bisk, Rowan Zellers, Ronan~Le Bras, Jianfeng Gao, and Yejin Choi.
\newblock Piqa: Reasoning about physical commonsense in natural language.
\newblock In \emph{Thirty-Fourth AAAI Conference on Artificial Intelligence},
  2020.

\bibitem[Clark et~al.(2018)Clark, Cowhey, Etzioni, Khot, Sabharwal, Schoenick,
  and Tafjord]{allenai:arc}
Peter Clark, Isaac Cowhey, Oren Etzioni, Tushar Khot, Ashish Sabharwal, Carissa
  Schoenick, and Oyvind Tafjord.
\newblock Think you have solved question answering? try arc, the ai2 reasoning
  challenge.
\newblock \emph{arXiv:1803.05457v1}, 2018.

\bibitem[Zellers et~al.(2019)Zellers, Holtzman, Bisk, Farhadi, and
  Choi]{zellers2019hellaswag}
Rowan Zellers, Ari Holtzman, Yonatan Bisk, Ali Farhadi, and Yejin Choi.
\newblock Hellaswag: Can a machine really finish your sentence?
\newblock In \emph{Proceedings of the 57th Annual Meeting of the Association
  for Computational Linguistics}, 2019.

\bibitem[Zhu et~al.(1997)Zhu, Byrd, Lu, and Nocedal]{L-BFGS-B}
Ciyou Zhu, Richard~H. Byrd, Peihuang Lu, and Jorge Nocedal.
\newblock Algorithm 778: L-bfgs-b: Fortran subroutines for large-scale
  bound-constrained optimization.
\newblock \emph{ACM Trans. Math. Softw.}, 23\penalty0 (4):\penalty0 550–560,
  December 1997.
\newblock ISSN 0098-3500.
\newblock \doi{10.1145/279232.279236}.
\newblock URL \url{https://doi.org/10.1145/279232.279236}.

\bibitem[Lee et~al.(2020)Lee, Perrone, Archambeau, and
  Seeger]{DBLP:journals/corr/abs-2003-10870}
Eric~Hans Lee, Valerio Perrone, C{\'{e}}dric Archambeau, and Matthias~W.
  Seeger.
\newblock Cost-aware bayesian optimization.
\newblock \emph{CoRR}, abs/2003.10870, 2020.
\newblock URL \url{https://arxiv.org/abs/2003.10870}.

\bibitem[Kingma and Ba(2014)]{Kingma2014AdamAM}
Diederik~P. Kingma and Jimmy Ba.
\newblock Adam: A method for stochastic optimization.
\newblock \emph{CoRR}, abs/1412.6980, 2014.
\newblock URL \url{https://api.semanticscholar.org/CorpusID:6628106}.

\bibitem[Lindauer et~al.(2022)Lindauer, Eggensperger, Feurer, Biedenkapp, Deng,
  Benjamins, Ruhkopf, Sass, and Hutter]{lindauer-jmlr22a}
Marius Lindauer, Katharina Eggensperger, Matthias Feurer, André Biedenkapp,
  Difan Deng, Carolin Benjamins, Tim Ruhkopf, René Sass, and Frank Hutter.
\newblock Smac3: A versatile bayesian optimization package for hyperparameter
  optimization.
\newblock \emph{Journal of Machine Learning Research}, 23\penalty0
  (54):\penalty0 1--9, 2022.
\newblock URL \url{http://jmlr.org/papers/v23/21-0888.html}.

\bibitem[Li et~al.(2018{\natexlab{b}})Li, Jamieson, DeSalvo, Rostamizadeh, and
  Talwalkar]{li2018hyperbandnovelbanditbasedapproach}
Lisha Li, Kevin Jamieson, Giulia DeSalvo, Afshin Rostamizadeh, and Ameet
  Talwalkar.
\newblock Hyperband: A novel bandit-based approach to hyperparameter
  optimization, 2018{\natexlab{b}}.
\newblock URL \url{https://arxiv.org/abs/1603.06560}.

\bibitem[Albalak et~al.(2023{\natexlab{b}})Albalak, Pan, Raffel, and
  Wang]{albalak2023efficientonlinedatamixing}
Alon Albalak, Liangming Pan, Colin Raffel, and William~Yang Wang.
\newblock Efficient online data mixing for language model pre-training,
  2023{\natexlab{b}}.
\newblock URL \url{https://arxiv.org/abs/2312.02406}.

\bibitem[Xie et~al.(2023)Xie, Pham, Dong, Du, Liu, Lu, Liang, Le, Ma, and
  Yu]{xie2023doremioptimizingdatamixtures}
Sang~Michael Xie, Hieu Pham, Xuanyi Dong, Nan Du, Hanxiao Liu, Yifeng Lu, Percy
  Liang, Quoc~V. Le, Tengyu Ma, and Adams~Wei Yu.
\newblock Doremi: Optimizing data mixtures speeds up language model
  pretraining, 2023.
\newblock URL \url{https://arxiv.org/abs/2305.10429}.

\bibitem[Ge et~al.(2025{\natexlab{b}})Ge, Ma, Chen, Li, and
  Ding]{ge2025bimixbivariatedatamixing}
Ce~Ge, Zhijian Ma, Daoyuan Chen, Yaliang Li, and Bolin Ding.
\newblock Bimix: A bivariate data mixing law for language model pretraining,
  2025{\natexlab{b}}.
\newblock URL \url{https://arxiv.org/abs/2405.14908}.

\bibitem[Goyal et~al.(2024{\natexlab{b}})Goyal, Maini, Lipton, Raghunathan, and
  Kolter]{goyal2024scalinglawsdatafiltering}
Sachin Goyal, Pratyush Maini, Zachary~C. Lipton, Aditi Raghunathan, and J.~Zico
  Kolter.
\newblock Scaling laws for data filtering -- data curation cannot be compute
  agnostic, 2024{\natexlab{b}}.
\newblock URL \url{https://arxiv.org/abs/2404.07177}.

\bibitem[Kaplan et~al.(2020)Kaplan, McCandlish, Henighan, Brown, Chess, Child,
  Gray, Radford, Wu, and Amodei]{kaplan2020scalinglawsneurallanguage}
Jared Kaplan, Sam McCandlish, Tom Henighan, Tom~B. Brown, Benjamin Chess, Rewon
  Child, Scott Gray, Alec Radford, Jeffrey Wu, and Dario Amodei.
\newblock Scaling laws for neural language models, 2020.
\newblock URL \url{https://arxiv.org/abs/2001.08361}.

\bibitem[Hoffmann et~al.(2022{\natexlab{b}})Hoffmann, Borgeaud, Mensch,
  Buchatskaya, Cai, Rutherford, de~Las~Casas, Hendricks, Welbl, Clark,
  Hennigan, Noland, Millican, van~den Driessche, Damoc, Guy, Osindero,
  Simonyan, Elsen, Rae, Vinyals, and
  Sifre]{hoffmann2022trainingcomputeoptimallargelanguage}
Jordan Hoffmann, Sebastian Borgeaud, Arthur Mensch, Elena Buchatskaya, Trevor
  Cai, Eliza Rutherford, Diego de~Las~Casas, Lisa~Anne Hendricks, Johannes
  Welbl, Aidan Clark, Tom Hennigan, Eric Noland, Katie Millican, George van~den
  Driessche, Bogdan Damoc, Aurelia Guy, Simon Osindero, Karen Simonyan, Erich
  Elsen, Jack~W. Rae, Oriol Vinyals, and Laurent Sifre.
\newblock Training compute-optimal large language models, 2022{\natexlab{b}}.
\newblock URL \url{https://arxiv.org/abs/2203.15556}.

\bibitem[Muennighoff et~al.(2023)Muennighoff, Rush, Barak, Scao, Piktus, Tazi,
  Pyysalo, Wolf, and
  Raffel]{muennighoff2023scalingdataconstrainedlanguagemodels}
Niklas Muennighoff, Alexander~M. Rush, Boaz Barak, Teven~Le Scao, Aleksandra
  Piktus, Nouamane Tazi, Sampo Pyysalo, Thomas Wolf, and Colin Raffel.
\newblock Scaling data-constrained language models, 2023.
\newblock URL \url{https://arxiv.org/abs/2305.16264}.

\bibitem[Ruan et~al.(2024)Ruan, Maddison, and
  Hashimoto]{ruan2024observationalscalinglawspredictability}
Yangjun Ruan, Chris~J. Maddison, and Tatsunori Hashimoto.
\newblock Observational scaling laws and the predictability of language model
  performance, 2024.
\newblock URL \url{https://arxiv.org/abs/2405.10938}.

\bibitem[Bergstra et~al.(2011)Bergstra, Bardenet, Bengio, and
  K\'{e}gl]{bergstra2011}
James Bergstra, R\'{e}mi Bardenet, Yoshua Bengio, and Bal\'{a}zs K\'{e}gl.
\newblock Algorithms for hyper-parameter optimization.
\newblock In \emph{Proceedings of the 25th International Conference on Neural
  Information Processing Systems}, NIPS'11, page 2546–2554, Red Hook, NY,
  USA, 2011. Curran Associates Inc.
\newblock ISBN 9781618395993.

\bibitem[Poloczek et~al.(2016)Poloczek, Wang, and
  Frazier]{poloczek2016multiinformationsourceoptimization}
Matthias Poloczek, Jialei Wang, and Peter~I. Frazier.
\newblock Multi-information source optimization, 2016.
\newblock URL \url{https://arxiv.org/abs/1603.00389}.

\bibitem[Wu et~al.(2019)Wu, Toscano{-}Palmerin, Frazier, and
  Wilson]{wu2019mfbo}
Jian Wu, Saul Toscano{-}Palmerin, Peter~I. Frazier, and Andrew~Gordon Wilson.
\newblock Practical multi-fidelity bayesian optimization for hyperparameter
  tuning.
\newblock \emph{CoRR}, abs/1903.04703, 2019.
\newblock URL \url{http://arxiv.org/abs/1903.04703}.

\end{thebibliography}

\newpage
\appendix
\section{Pretraining Runs Details}\label{appendix:pretraining}

We use the OLMo 2 \cite{olmo20242olmo2furious} package for training our language models. 
The model configurations are

\begin{table}[h]
    \centering
    \begin{tabular}{lcccc}
        \toprule
        \textbf{Group} & \textbf{d\_model} & \textbf{n\_heads} & \textbf{n\_layers} & \textbf{Runs} \\
        \midrule
        20M  & 256   & 8  & 8  & 115 \\
        60M  & 512   & 8  & 8  & 71  \\
        150M & 768   & 12 & 12 & 53  \\
        300M & 1024  & 16 & 16 & 74  \\
        500M & 1280  & 16 & 16 & 39  \\
        700M & 1536  & 16 & 16 & 52  \\
        1B   & 2048  & 16 & 16 & 68  \\
        \bottomrule
    \end{tabular}
    \caption{Model Architecture Details by Group with Number of Runs}
\end{table}

To decide the training mixtures of the experiments, for each experiment, we randomly sample from the probability simplex using the Dirichlet distribution of order $n$ (number of training datasets), where we use the parameters $\alpha_i = 1, \forall i \in [n]$.
Other training configurations (learning rate, momentum, etc.) are directly taken from OLMo's configuration files (e.g. \href{https://github.com/allenai/OLMo/blob/main/configs/tiny/OLMo-700M.yaml}{700M})
We study the compute optimal regime \cite{Hoffmann_Chinchilla}: for each 1B model run, we used 20B tokens in total for training.
In the interest of collecting more runs, all other model scales are trained on 10B tokens. 

\section{Bayesian Optimization Details}\label{appendix:bopt}
For MFMS-GP, the cost of evaluating a run at a particular model scale is taken from the number of FLOPS the corresponding model scale costs during the pretraining runs. 
The costs are scaled appropriately such that a unit of cost corresponds FLOPS required to train 1B model for 1 training step. 

Additionally, since it is prohibitively expensive to optimize EI for each of 19700 training steps, for multi-fidelity multi-scale GP, we limit the space of training steps to be $\mathcal Z = \{6000, 12000, 19700\}$. 

The GP hyperparameters are trained using the Adam optimizer with a 0.1 learning rate for 50 iterations. 

As mentioned in Section~\ref{subsec:mfms-gp}, we optimize EI within $(m, z)$ tuple.
For this optimization, we initiate 5 random probability weights drawn from the Dirichlet distribution and perform a gradient search for greater EI over the probability simplex. 

Occasionally, the GP would be too certain of its posterior prediction such that the optimized EIs are all small in magnitude. 
Therefore, when the optimal EI is below a certain threshold, we lower the length scales of the RBF kernels to encourage more exploration. 
The threshold is set to be $1e^{-4}$, and the length scales would be lowered to $95\%$ of their original values.

As a measure to encourage selecting higher cost evaluations later in the optimization cycle, instead of using $\text{EI}_{\text{pu}}(\lambda) = \frac{\text{EI}(\lambda)}{c(m, z)}$, we introduce an additional parameter $\alpha$ that controls the importance of cost, and pick the configuration that maximizes $\frac{\text{EI}(\lambda)}{c(m, z)^\alpha}$. 
Initially $\alpha=1$, and it decays by $1\%$ for every step of the Bayesian optimization. 
\section{Kernel Comparisons}\label{appendix:kernel}
The kernel function $k(x, x')$ in a GP defines the covariance between different input points. 
In the main results, we use a RBF kernel $k(\boldsymbol{w}, \boldsymbol{w}') = \text{exp}(-\frac{d(\boldsymbol{w}, \boldsymbol{w}')}{2\sigma^2})$ for the data proportions $\boldsymbol{w}$, where $d(\boldsymbol{w}, \boldsymbol{w}') = \|\boldsymbol{w} - \boldsymbol{w}'\|^2$ is the squared $L^2$ distance between the two probabilities. 
In this appendix, we experiment with distance metrics that may be more suitable for probabilities. 

\begin{figure*}[!th]
    \centering
    \includegraphics[width=\textwidth]{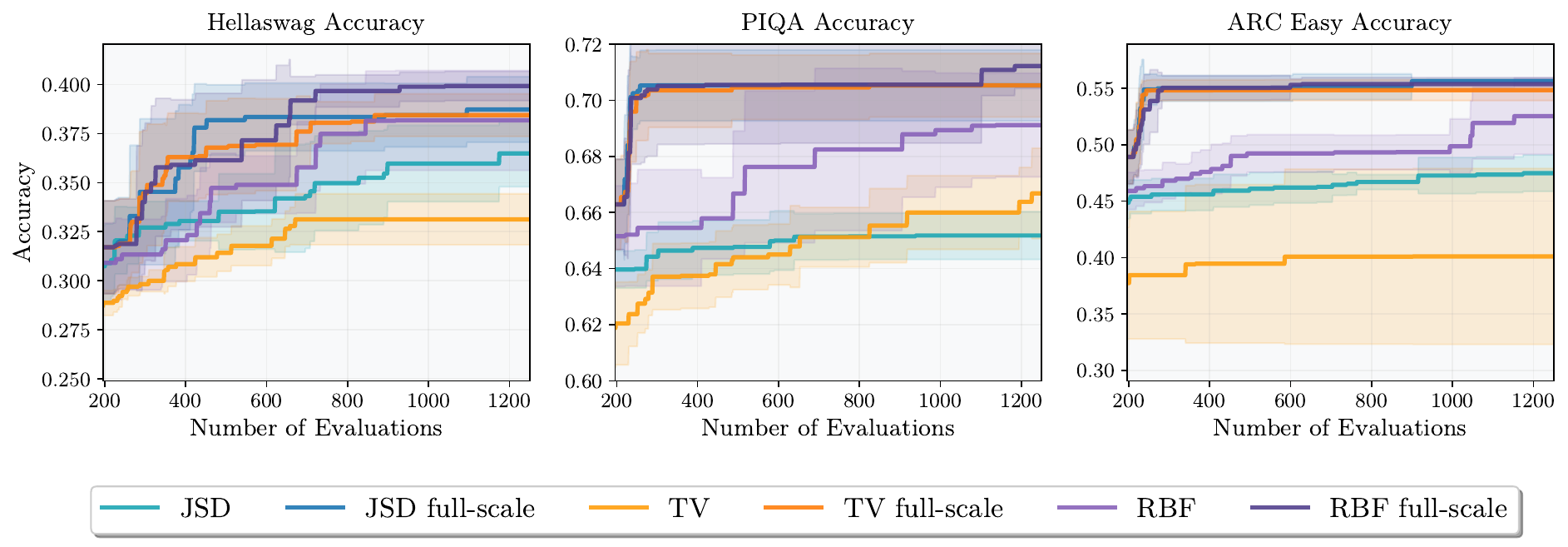}
    \caption{Comparing different Gaussian process kernels on maximizing accuracy in the downstream tasks.}
    \label{fig:kernel1}
\end{figure*}

\begin{figure*}[!th]
    \centering
    \includegraphics[width=\textwidth]{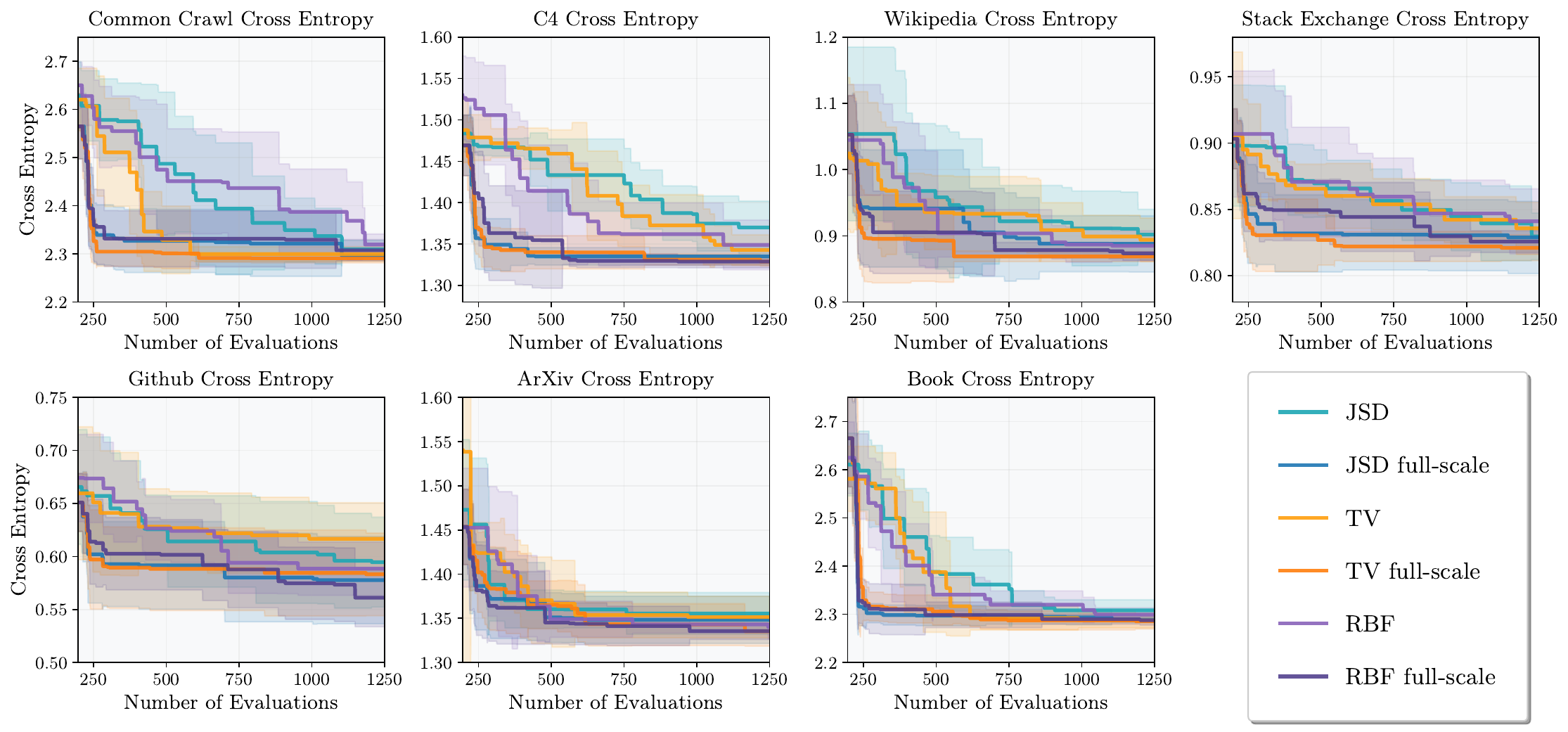}
    \caption{Comparing different Gaussian process kernels on minimizing the validation cross-entropy losses}
    \label{fig:kernel2}
\end{figure*}
Specifically, we consider the Total Variance (TV) distance and the Jensen–Shannon divergence (JSD), a symmetric Kullback–Leibler divergence, for the data proportions: 
\begin{eqnarray}
    \text{TV}: && d(\boldsymbol{w}, \boldsymbol{w}') = \|\boldsymbol{w} - \boldsymbol{w}'\|_1
    \nonumber
    \\ 
    \text{JSD}: && d(\boldsymbol{w}, \boldsymbol{w}') = 
        \frac{1}{2} \text{KL} \left(
            \boldsymbol{w}\  \| \ \boldsymbol{\bar w}
        \right)
        + \frac{1}{2} \text{KL} \left(
            \boldsymbol{w}'\  \| \ \boldsymbol{\bar w}
        \right)
    \nonumber
\end{eqnarray}
where $\text{KL}(\cdot)$ denotes the Kullback–Leibler divergence, and $\boldsymbol{\bar w} = \frac{\boldsymbol{w} + \boldsymbol{w}'}{2}$. 

We found that on optimizing accuracy, the simple RBF kernels generally perform better. 
However, occasionally, JSD (on HellaSwag of Figure.~\ref{fig:kernel1}) or TV (on Common Crawl, Wikipedia, and Stack Exchange of Figure.~\ref{fig:kernel2}) yield better results.

\end{document}